%% file: main.tex
\documentclass[sigconf]{acmart}

\AtBeginDocument{%
  \providecommand\BibTeX{{%
    \normalfont B\kern-0.5em{\scshape i\kern-0.25em b}\kern-0.8em\TeX}}}

\copyrightyear{2021}
\acmYear{2021}
\setcopyright{acmlicensed}\acmConference[WSDM '21]{Proceedings of the Fourteenth ACM International Conference on Web Search and Data Mining}{March 8--12, 2021}{Virtual Event, Israel}
\acmBooktitle{Proceedings of the Fourteenth ACM International Conference on Web Search and Data Mining (WSDM '21), March 8--12, 2021, Virtual Event, Israel}
\acmPrice{15.00}
\acmDOI{10.1145/3437963.3441736} \acmISBN{978-1-4503-8297-7/21/03}

\usepackage[utf8]{inputenc} % allow utf-8 input
\usepackage[T1]{fontenc}    % use 8-bit T1 fonts
\usepackage{hyperref}       % hyperlinks
\usepackage{url}            % simple URL typesetting
\usepackage{booktabs}       % professional-quality tables
\usepackage{amsfonts}       % blackboard math symbols
\usepackage{nicefrac}       % compact symbols for 1/2, etc.
\usepackage{microtype}      % microtypography
\usepackage{graphicx}
\usepackage{amsmath,amsfonts,amsthm}
\usepackage{array}
\usepackage{subcaption}
\usepackage{bbm}
\usepackage{bm}
\usepackage{multicol}
\usepackage{lipsum}
\usepackage{wrapfig}
\usepackage[ruled]{algorithm2e}
\usepackage{arydshln}
\usepackage{multirow}
\usepackage{balance}

\newtheorem{claim}{Claim}

\newtheorem{lemma}{Lemma}
\newtheorem{remark}{Remark}
\newtheorem{theorem}{Theorem}

\newcommand{\ind}{\mathbbm{1}}
\newcommand{\diff}{\text{d}}
\newcommand{\diag}{\text{diag}}
\newcommand{\Ical}{\mathcal{I}}
\newcommand{\Ncal}{\mathcal{N}}
\newcommand{\Dcal}{\mathcal{D}}
\newcommand{\Lcal}{\mathcal{L}}

\newcommand{\Ebb}{\mathbb{E}}
\newcommand{\Rbb}{\mathbb{R}}
\newcommand{\Pbb}{\mathbb{P}}
\newcommand{\ybf}{\mathbf{y}}
\newcommand{\Xbf}{\mathbf{X}}
\newcommand{\xbf}{\mathbf{x}}

\newcommand{\Zbf}{\mathbf{Z}}
\newcommand{\zbf}{\mathbf{z}}
\newcommand{\wbf}{\mathbf{w}}
\newcommand{\Rbf}{\mathbf{R}}
\newcommand{\Ubf}{\mathbf{U}}

\newcommand{\Vbf}{\mathbf{V}}

\newcommand{\Sigmabf}{\boldsymbol \Sigma}
\newcommand{\thetabf}{\boldsymbol \theta}
\newcommand{\epsilonbf}{\boldsymbol \epsilon}

\begin{document}

\fancyhead{}
\title{Theoretical Understandings of Product Embedding for E-commerce Machine Learning}


\author{Da Xu}
\affiliation{%
  \institution{Walmart Labs}
  \city{Sunnyvale}
  \state{California}
  \country{USA}
}
\email{DaXu5180@gmail.com}

\author{Chuanwei Ruan}
\authornote{Work was done when the author was with Walmart Labs.}
\affiliation{%
  \institution{Instacart}
  \city{San Francisco}
  \state{California}
  \country{USA}
}
\email{RuanChuanwei@gmail.com}

\author{Evren Korpeoglu \\ Sushant Kumar \\ Kannan Achan}
\affiliation{%
  \institution{Walmart Labs}
  \city{Sunnyvale}
  \state{California}
  \country{USA}
}
\email{[EKorpeoglu,SKumar4,KAchan]}
\email{@walmartlabs.com}

\begin{abstract}
Product embeddings have been heavily investigated in the past few years, serving as the cornerstone for a broad range of machine learning applications in e-commerce. Despite the empirical success of product embeddings, little is known on how and why they work from the theoretical standpoint. Analogous results from the natural language processing (NLP) often rely on domain-specific properties that are not transferable to the e-commerce setting, and the downstream tasks often focus on different aspects of the embeddings. We take an e-commerce-oriented view of the product embeddings and reveal a complete theoretical view from both the representation learning and the learning theory perspective. We prove that product embeddings trained by the widely-adopted skip-gram negative sampling algorithm and its variants are sufficient dimension reduction regarding a critical product relatedness measure. The generalization performance in the downstream machine learning task is controlled by the alignment between the embeddings and the product relatedness measure. Following the theoretical discoveries, we conduct exploratory experiments that supports our theoretical insights for the product embeddings.
\end{abstract}

\begin{CCSXML}
<ccs2012>
<concept>
<concept_id>10002950.10003648</concept_id>
<concept_desc>Mathematics of computing~Probability and statistics</concept_desc>
<concept_significance>500</concept_significance>
</concept>
<concept>
<concept_id>10002951.10003317</concept_id>
<concept_desc>Information systems~Information retrieval</concept_desc>
<concept_significance>500</concept_significance>
</concept>
<concept>
<concept_id>10003752.10010070.10010071</concept_id>
<concept_desc>Theory of computation~Machine learning theory</concept_desc>
<concept_significance>500</concept_significance>
</concept>
</ccs2012>
\end{CCSXML}

\ccsdesc[500]{Mathematics of computing~Probability and statistics}
\ccsdesc[500]{Information systems~Information retrieval}
\ccsdesc[500]{Theory of computation~Machine learning theory}
\keywords{Representation learning; Product relation; Information theory; Sufficient dimension reduction; Machine learning theory}


\maketitle

\section{Introduction}
\label{sec:introduction}
\input{introduction}

\section{Background and Related Work}
\label{sec:background}
\input{background}

\section{Nonlinear Projection and Sufficient Dimension Reduction}
\label{sec:SDR}
\input{SDR}

\section{Properties of the Product Relatedness Measure}
\label{sec:PMI}
\input{PMI}

\section{The Generalization Bound of Product Embeddings}
\label{sec:generalization}
\input{generalization}

\section{Experiments and Results}
\label{sec:experiment}
\input{experiment}

% \section{Conclusion}

% \appendix 
% \section{Proofs}
% The proofs are provided in the online material: {\footnotesize
% \url{https://github.com/StatsDLMathsRecomSys/Theoretical-Understandings-of-Product-Embedding}}.
% \section{Proofs}
% \input{appendix}

\bibliographystyle{ACM-Reference-Format}
\balance
\bibliography{references}


% \twocolumn[
%   \begin{@twocolumnfalse}
%   \appendix
%     \section{Proofs}
%     \input{proofs}
%   \end{@twocolumnfalse}
% ]
\onecolumn
  \appendix
    \section{Proofs}
    \input{proofs}

\end{document}

%% file: introduction.tex
Model interpretation and understanding play a critical role in e-commerce machine learning. Unlike the other domains where deep learning algorithms are being favored regardless of their black-box nature, model interpretability is often equally important as empirical performance in e-commerce, due to its closer connections with business and customers, as well as a more profound impact on revenue and social accountability including privacy, security and fairness \cite{pennanen2006trust,yazdanifard2011security}. The underlying theoretical properties often justify model interpretation. 
Without sufficient theoretical understandings, model developers need to rely on intuitions and unverifiable assumptions to explain the inductive bias rather than providing reliable theoretical support and guarantee, which can easily result in mismatches between the purpose of model design and the actual working mechanism. It poses severe challenges on model diagnostics, which is an indispensable part of any industrial deployment. 
On the other hand, sacrificing the interpretability for a higher model complexity may not lead to better empirical performance. Several recent papers have challenged the state-of-the-art deep learning recommendation algorithms against the vanilla collaborative filtering, ending up finding worse performances from deep learning on various benchmark datasets \cite{dacrema2019we,rendle2020neural}. 
All the above concerns motivate our exploration of the theoretical perspective of product embeddings - the cornerstone for a considerable amount of machine learning models in e-commerce \cite{barkan2016item2vec,grbovic2015commerce,vasile2016meta,xu2020knowledge,wang2018billion,xu2020product}.

Modern e-commerce machine learning favors the embedding models over classical feature-based approaches because of their computation efficiency and compatibility with more model architectures. Training product embeddings using the skip-gram negative sampling (SGNS) algorithm and its variants are highly efficient and scalable, even for billions of items and records \cite{mikolov2013efficient,mikolov2013distributed,wang2018billion}. A handful of open-source subroutines are available for easy implementation and modification under problem-specific needs \cite{goldberg2014word2vec}. 
By treating the product embeddings as vectors (usually of several hundred dimensions) encoded with useful product information, the computations of downstream tasks are simplified after converted to the low-dimensional Euclidean space \cite{huang2020embedding}. 
Replacing the product features with embedding in the downstream tasks significantly enriches the candidate models and reduces the feature engineering costs \cite{cheng2016wide,covington2016deep}. The incompatibility issue is common for industrial problems, since the product features are often mixtures of quantitative and categorical variables that expand a huge irregular space that few modern machine learning models are suitable. 
Different approaches have been proposed to train product embeddings using SGNS with problem-specific modification, including \textsl{Item2vec} \cite{barkan2016item2vec}, \textsl{Prod2vec} \cite{grbovic2015commerce,vasile2016meta}, \textsl{Triple2vec} \cite{wan2018representing}, \textsl{MetaPath2vec} \cite{dong2017metapath2vec}, \textsl{CompProd2vec} (complementary product embedding) \cite{xu2020knowledge} and product knowledge graph \cite{xu2020product}. Despite the different formulations of the input data structure and regularization, the core component remains to be the SGNS induced by the input co-occurrence statistic (see Section \ref{sec:background} for details). The promising results from industrial applications also support the progress in the academic research on product embedding. Some of the papers have reported successful deployments, highlighting product embeddings as part of the mature solution for various online services.

Nevertheless, our understanding of product embedding is still inadequate to the classical models such as collaborative filtering and factorization machines \cite{rendle2010factorization,schafer2007collaborative,sarwar2001item}. Despite the conjectures and claims that product embedding encodes the useful features and relations, research in this domain has yet found an exact mapping to unveil how product embedding captures the signals and why they are useful for downstream tasks. The recent progress 
for the model understanding in the NLP domain, though highlighting certain aspects of the SGNS algorithm \cite{levy2014neural,allen2019analogies,cotterell2017explaining,arora2016latent,stratos2015model}, does not directly transfer to the e-commerce setting due to different emphasis and data assumptions. We provide in-depth discussions in Section \ref{sec:background}. In general, the product embedding is only partially understood, and the remaining unknown factors may still raise concerns from time to time and set barricades for the more thorough analysis and further improvements.

Our work is dedicated to providing an advanced theoretical understanding of product embedding for e-commerce machine learning. The first key result establishes the equivalence between training product embedding and finding the sufficient dimension reduction \cite{globerson2003sufficient} of a \emph{product relatedness measure} induced by the \emph{co-occurrence statistic}. The product embedding, as a consequence, is optimal in an information-theoretical perspective. We then highlight several properties of the product relatedness measure, including its finite-sample tail bound and several domain-specific functionalities. The second key result shows the generalization bound for using product embedding in downstream machine learning tasks. It turns out that the generalization error is controlled by the alignment between the spectral spaces of product embedding and the product relatedness measure. In summary, we provide advanced theoretical understandings by answering: 1. what data distribution is product embedding representing; 2. how product embedding is representing the signal in e-commerce data; 3. why product embedding is useful for downstream tasks.

The practical implications of our results are two folds. Firstly, since product embedding is an (information-theoretical) optimal dimension reduction of the problem-specific product relatedness measure, its quality (meaningfulness) depends on the relatedness measure, which can be examined using the tail bound in Section \ref{sec:PMI}. Secondly, the applicability (usefulness) of product embedding in downstream tasks can be examined in advance by checking how well they reconstruct the eigenspace of the product relatedness measure. They provide further understandings and some guidelines for obtaining more meaningful and useful product embedding, which we give a thorough exploration in our experiments. To the best of our knowledge, our paper provides the first advanced theoretical understanding of product embedding, and we conclude our contributions as follow.
\begin{itemize}
    \item We establish the equivalence between training product embedding and sufficient dimension reduction with respect to the product relatedness measure.
    \item We provide the finite-sample tail bound and verify several domain-specific functionalities for the relatedness measure.
    \item We give a generalization bound for using product embedding in downstream tasks, and further illustrate our theoretical arguments via experiments.
\end{itemize}

%% file: background.tex
By convention, we use uppercase letters to denote random variables, lowercase letters to denote observations and scalars, bold-font letters to denote vectors and matrices, $D_{KL}(p\, \|\, q)$ to denote the Kullback-Leibler divergence between distribution $P$ and $Q$ with the corresponding density function $p$, $q$. 
% The main notations are introduced in Table \ref{tab:notations}.

\begin{table}[htb]
    \centering
    \begin{tabular}{|m{2cm}|m{5.6cm}|}
    \hline
        $\Ical$, $\text{Neg}(\Ical)$ & The set of all products, and the negative samples drawn from $\Ical$.  \\ \hline
        $\Ncal(i)$ & The neighborhood set for product $i\in\Ical$. \\ \hline
        $\Dcal(\Ncal)$ & The data generating mechanism with respect to the input data structure as well as the definition of neighborhood $\Ncal(.)$. See Figure \ref{fig:illustration} for examples. We omit the dependency on $\Ncal(.)$ for notation simplicity when no confusion arises. \\ \hline
        $P_i(\Dcal)$, $P_{i,j}(\Dcal)$ & The marginal frequency of product $i$ and product pair $i,j$ for $i,j \in \Ical$, with respect to the data generating mechanism $\Dcal$. \\ \hline
        $n$, $N_{i}(\Dcal)$, $N_{i,j}(\Dcal)$ & The total number of records, and the (co-)occurrence statistics such that $P_i(\Dcal) = N_{i}(\Dcal) / n$ and $P_{i,j}(\Dcal) = N_{i,j}(\Dcal) / n$. \\ \hline
        $\zbf_i$, $\tilde{\zbf}_i \in \Rbb^d$ & The two embeddings for product $i\in \Ical$. The additional embedding helps handling the asymmetric product relations such that $\langle \zbf_i, \tilde{\zbf}_j \rangle \neq \langle \zbf_j, \tilde{\zbf}_i \rangle$. \\ \hline
        $p(O=1 | i, j)$ & The probability that product $i$ and $j$ co-occur in the same neighborhood, i.e. $p\big(1 [j \in \Ncal(i)])$. \\
        \hline
    \end{tabular}
    \caption{\small Notations. Notice that $P_i(\Dcal)$, $P_{i,j}(\Dcal)$, $N_{i}(\Dcal)$ and $N_{i,j}(\Dcal)$ are random variables with their stochsticity induced by the data generating mechanism $\Dcal(\Ncal)$.}
    \label{tab:notations}
\end{table}

\begin{figure}
    \centering
    \includegraphics[width=0.7\linewidth]{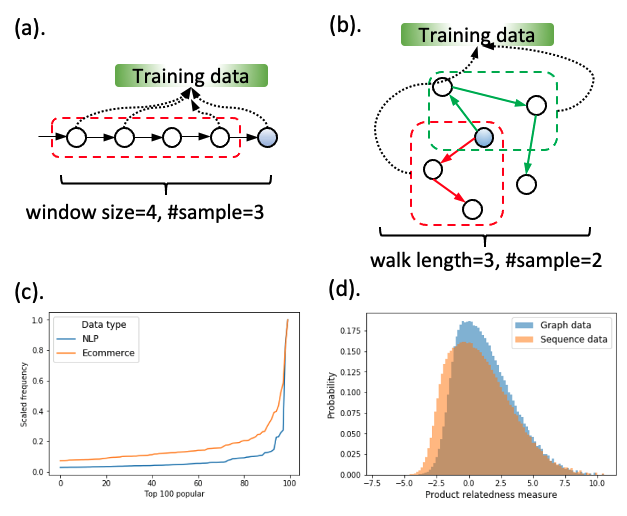}
    \caption{\small (a), (b): Visual example of generating training data for SGNS using sequence and graph structure. The sampling steps induce the randomness in training data. (c). The scaled frequency for the 100 most occurred words/products in an NLP corpus and a pubic e-commerce data (see Section \ref{sec:experiment}). Observe that the overlap only occurs for the top few cases. (d). The empirical distribution of the product relatedness measure $R_{i,j}$, for both the sequence-structured and graph-structured training data (as shown in (a) and (b)), generated from the pubic e-commerce data. }
    \label{fig:illustration}
\end{figure}

Product embeddings are trained on the input data that structured specifically to reflect particular product relations. 
% We provide sketched visual illustrations in Figure ??? to demonstrate several (simplified) data structures proposed in previous literature.
The \textsl{skip-gram negative sampling} algorithm optimizes the embeddings to capture the desired product relation via inner products \cite{mikolov2013efficient,mikolov2013distributed}. Product pairs from the same neighborhood are likely to have closer relations and are hence treated as positive samples. A fundamental difference between learning product embedding and word embedding is that the notion of neighborhood can be random variables in the e-commerce setting. The stochasticity in neighborhood may be induced by its own definition, e.g. the outcome of random walks (\textsl{MetaPath2vec} \cite{dong2017metapath2vec}, \textsl{ProdNode2vec} \cite{wang2018billion}), or by the follow-up sampling steps, e.g. sampling the products from a given context window (\textsl{CompProd2vec} \cite{xu2020knowledge}).
Therefore, the realization of neighborhood varies for different input data structures and problem settings (Figure \ref{fig:illustration}), and the resulting co-occurrence statistics are random variables as well. To better notate the dependency on the underlying data generating mechanism as well as the neighborhood definition, we design our notation system as shown in Table \ref{tab:notations}. 

As an unsupervised learning approach, the SGNS objective function is designed to characterize the contrasts between positive and negative samples using the sigmoid function:
\begin{equation}
\label{eqn:setup}
\begin{split}
    & p(O=1 | i, j) = \sigma(\zbf_i^{\intercal}\tilde{\zbf}_j) =  \frac{1}{1+\exp(-\zbf_i^{\intercal}\tilde{\zbf}_j)}, \\
    & p(O=0 | i, j) = \sigma(-\zbf_i^{\intercal}\tilde{\zbf}_j) = \frac{1}{1+\exp(\zbf_i^{\intercal}\tilde{\zbf}_j)}, \\
    \ell_{i,j} &= -\log p(O=1 | i, j) + k\cdot \Ebb_{k\sim \text{Neg}(\Ical)} \big[\log  p(O=0 | i, k) \big],
\end{split}
\end{equation}
here $k$ is the number of negative samples.
Without loss of generality, the modified objective functions for different product embedding algorithms can be summarized by:
\begin{equation}
\label{eqn:emb-loss}
\ell = \sum_{i, j \in \Ncal(i)} \ell_{i,j} + \text{Reg}(\Zbf, \tilde{\Zbf}), \, \Zbf=[\zbf_1,\ldots,\zbf_{|\Ical|}], \, \tilde{\Zbf}=[\tilde{\zbf}_1,\ldots,\tilde{\zbf}_{|\Ical|}],
\end{equation}
where $\text{Reg}(\Zbf, \tilde{\Zbf})$ is the (implicit) problem-specific regularization on the product embeddings \footnote{The original papers do not introduce their objectives by this form, and it is possible that an explicit expression for the regularizer is nonexistent. However, by viewing the algorithms in this way, we distinguish the core component and the problem-specific structures, which is necessary for analytical purposes. }, e.g. products with similar features should have closer embeddings, or the regularization takes account of the user bias whenever user id is presented in the data. Set aside the regularizations, the NLP community has made considerable efforts to shed insights to the SGNS objective function \cite{levy2014neural,allen2019analogies,cotterell2017explaining,arora2016latent,stratos2015model}. Although the skip-gram model has been well understood \cite{guthrie2006closer}, the negative sampling has changed the algorithm fundamentally. The major breakthrough was made in \cite{levy2014neural}, showing that the optimal embedding is given by the point-wise mutual information (PMI) matrix. However, their result requires a full rank assumption on the embedding matrix, i.e. the embedding dimension is larger than the number of entities, which is not practical. Other papers that build direct connections between PMI and embedding either require a specific type of entity distribution or a underlying generative model from embeddings \cite{cotterell2017explaining,arora2016latent,stratos2015model}, which are not reasonable for the e-commerce setting. Two recent papers have provided probabilistic interpretation for embeddings trained by SGNS \cite{allen2019analogies,allen2019vec}, nonetheless, they focus on the compositional properties of trained embeddings rather than the training process and downstream tasks. 

The notion of sufficient dimension reduction is inspired by the sufficient statistics of exponential family. Broadly speaking, when predicting $Y$ with $X$, if all the information in $X$ about $Y$ can be compressed into a dimension reduction $f(X)$, i.e. $Y \perp X \, |\, f(X)$ or $p(Y|X)$ and $p\big(Y|f(X)\big)$ are the same, then $f(X)$ is a sufficient dimension reduction (SDR). Finding the SDR $f(X)$ is equivalent to solving a constraint minimax optimization \cite{globerson2003sufficient}:
\begin{equation}
    \max_{f(X)} \min_{\substack{q(y,x): \\ \int_x q(y,x)\diff x = p(y),\, \int_y q(y,x)\diff y = p(x), \\ \Ebb_{q(x|y)} [f(X)] = \Ebb_{p(x|y)} [f(X)]}} I\big(q(Y, X)\big),
\end{equation}
where $I\big(q(Y, X)\big)$ is the Shannon mutual information, and $p(x)$, $p(y)$ corresponds to the marginal distribution of $p(x,y)$. Briefly put, the above minimax problem introduces a proxy distribution $q(y, x)$ induced by the SDR $f(X)$ (reflected by the third constraint under minimization) while maintaining the marginal distribution of $X$ and $Y$ (see the first two constraints under minimization). The SDR $f(X)$ is then optimized under the "maximum entropy principle", by maximizing the Shannon mutual information of $q(y, x)$. Intuitively speaking, SDR is finding the dimension reduction that compresses the maximum amount of information while still agreeing with the observed data distribution. Applying the variational principle and strong duality arguments, it has been shown in \cite{globerson2003sufficient} that the dual problem is given by:
\begin{equation}
\label{eqn:sdr-def}
    \min_{q\big(y, f(x)\big)} D_{KL}\big( p(Y,X) \, \| \, q(Y,f(X))   \big).
\end{equation}
Sufficient dimension reduction is a powerful tool to examine the information-theoretical optimality of compressed representation, which we prove for the product embedding in the following section.

%% file: SDR.tex
Formally, apart from the embedding regularization terms in (\ref{eqn:emb-loss}), the objective function of SGNS for training product embedding has the alternative expression by aggregating the contribution to the positive and negative samples from each item pair:
\begin{equation}
\label{eqn:sgns-loss}
    \ell(\Dcal) = \sum_{i, j \in \Ical} N_{i,j}(\Dcal) \log \sigma(\zbf_i^{\intercal}\tilde{\zbf}_j) + \frac{k}{n} N_i(\Dcal) N_j(\Dcal) \log \sigma(-\zbf_i^{\intercal}\tilde{\zbf}_j).
\end{equation}
Note that $\ell_{\text{SGNS}}$ is a random variable because it conditions on the unknown data generating mechanism $\Dcal$. When the input data, its structure and the definition of neighborhood are given, i.e. $\Dcal$ is realized, the loss function is then fixed and observed. The loss function may seem peculiar at the first glance, but it is a special instance of nonlinear projection of a product relatedness measure:
\begin{equation}
\label{eqn:pmi}
R_{i,j} = \log \big\{n N_{i,j}(\Dcal) /\big( N_i(\Dcal)N_j(\Dcal) \big) \big\}. 
\end{equation}
Taking the gradient of $\ell(\Dcal)$ with respect to $\zbf_i$ (the role of $\zbf_i$ and $\tilde{\zbf}_i$ is symmetric so we may consider either one), we obtain:
\begin{equation}
\label{eqn:grad-sgns}
\nabla_{\zbf_i} \ell(\Dcal) = \tilde{Z} \diag(\wbf_i) \underbrace{\Big\{\sigma\big([R_{i,1}, \ldots, R_{i,|\Ical|}]\big) - \sigma(\langle \zbf_i, \tilde{\Zbf}\rangle) \Big\}}_{\text{error term}},
\end{equation}
where $\diag(\wbf_i)$ is the diagonal weight matrix with each instance is by given $\wbf_{i,j} = p_{i,j}(\Dcal) + kp_{i}(\Dcal)p_{j}(\Dcal)$. Bearing (\ref{eqn:grad-sgns}) in mind, let us revisit the weighted least square matrix factorization for $\big[R_{i,j}\big]_{i,j=1}^{|\Ical|}$ where the weights are also given by $\wbf_{i,j}$: $
\ell_{\text{ls}}(\Dcal) = \sum_{i,j} \wbf_{i,j} \big(R_{i,j} - \zbf_i^{\intercal}\tilde{\zbf}_j \big)^2
$. The gradient with respect to $\zbf_i$ is given by:
\begin{equation}
\label{eqn:grad-ls}
\nabla_{\zbf_i} \ell_{\text{ls}}(\Dcal) = \tilde{Z} \diag(\wbf_i) \underbrace{\Big\{[R_{i,1}, \ldots, R_{i,|\Ical|}] - \langle \zbf_i, \tilde{\Zbf}\rangle \Big\}}_{\text{error term}}.
\end{equation}
Comparing (\ref{eqn:grad-ls}) with (\ref{eqn:grad-sgns}), we see that the only difference lies in the error term which measures the deviation of projecting $\zbf_i$ to the space of $\tilde{\Zbf}$ and the target vector $[R_{i,1}, \ldots, R_{i,|\Ical|}]$:
\begin{itemize}
    \item for the SGNS gradient in (\ref{eqn:grad-sgns}), the error is measured on the nonlinear space induced by the $\sigma(\cdot)$ transformation;
    \item for the lest-square objective in (\ref{eqn:grad-ls}), the error is measured on the regular Euclidean space.
\end{itemize}
By comparing with the least-square matrix factorization, which characterizes linear project by the definition, we see that project embeddings trained by SGNS algorithm are essentially non-linear projections of the product relatedness matrix $\Big[R_{i,j}\Big]_{i,j=1}^{|\Ical|}$. 

\begin{remark}[Product relatedness measure and point-wise mutual information]
The point-wise mutual information (PMI) is defined by $\log \big\{ p(i,j) / p(i)p(j) \big \}$, which is a data-specific measurement on how much information about entity $j$ is in entity $i$. Notice that the product relatedness measure is a random variable while the PMI is a fixed quantity. Upon a realization of the data generating mechanism $\Dcal$, the PMI$_{i,j}$ computed by the observed data is an estimation of the the product relatedness measure $R_{i,j}$.
\end{remark}

Upon realizing that product embeddings are nonlinear projections, the next question is in what sense is the particular nonlinear projection optimal. Unlike linear projection whose optimality in the $\ell_2$ norm is well understood, there is no rule-of-thumb method to analyze nonlinear projections. However, our exploration from the SDR perspective is not by mere guessing. The key intuition is from the result in \cite{levy2014neural}, that if there are no constraints on the embedding dimension, then the minimizer of the SGNS objective is exactly given by the corresponding product relatedness measure $\Big[R_{i,j}\Big]_{i,j=1}^{|\Ical|}$: $\big[R_{i,1}, \ldots, R_{i,|\Ical|}\big] = \arg \min_{\zbf_i} \ell(\Dcal)$, $i=1,\ldots,|\Ical|$. 

With a pre-specified dimension $d < |\Ical|$, the objective becomes: 
$\arg \min_{\zbf_i \in \Rbb^d} \ell(\Dcal)$,
which constraints the solution to the convex subspace of $\Rbb^d$. The convexity of the constraint space often leads to nice relations between the unconstrained optimum and constrained optimum, e.g. the maximum-likelihood estimation leads to the locally optimal instance (in the constraint model space) in terms of the \emph{KL divergence}. The SGNS objective, with scrutiny, is also maximizing a particular likelihood function. We formalize the above intuition in the following claim.

\begin{claim}
\label{claim:sdr}
Let $q\big(O \, \big| \, \Dcal;\, \Zbf,\tilde{\Zbf}\big)$ be the co-occurrence probability computed by the embedding as in (\ref{eqn:setup}). At global optimum, the embedding matrices are given by the product relatedness matrix that gives the co-occurrence probability $p\big(O \, \big| \, \Dcal;\, R\big)$. The minimizer of the SGNS objective function is characterized by:
\begin{equation}
\label{eqn:sgns-sdr}
\underset{\Zbf, \tilde{\Zbf} \in \Rbb^d}{\text{minimize}} \, \, D_{KL}\Big( q\big(O \, \big| \, \Dcal;\, \Zbf,\tilde{\Zbf}\big) \, \big\| \, p\big(O \, \big| \, \Dcal;\, R\big) \Big).
\end{equation}
According to (\ref{eqn:sdr-def}), the product embedding is the sufficient dimension reduction of product relatedness measure with respect to the co-occurrence probability.
\end{claim}
The proof is provided in the appendix.
Much of the analysis in this section holds for the general embedding settings as well. However, researchers from other domains, e.g. the NLP community, do not take the same approach. We explicitly consider the uncertainty from data generating mechanism, which is necessary for product embeddings because the input data structures and definitions of neighborhood are very different under various problems. On the other hand, the NLP community treats the data as fixed and given by the corpus, and the definition of neighborhood often has less impact on the outcome. In contrast, our particular interests in this type of results are driven by the pursuit of model interpretability for e-commerce machine learning.

Recognizing the product embedding as nonlinear projection plus sufficient dimension reduction of the product relatedness measure for the co-occurrence probability provides understandings of their nature. It is also an essential step towards analyzing their domain-specific theoretical properties, i.e. their meaningfulness and usefulness in the e-commerce setting, which are the topics of the next two sections. 

%% file: PMI.tex
As a result of Claim \ref{claim:sdr}, the understandings of the product relatedness measure can be transferred to the product embedding since it is the information-theoretical optimal compression in the $\Rbb^d$ space. We explore two types of domain-specific properties for the product relatedness measure. The first property is concerned with the false association problem of product relation, which leads to a practical data cleaning procedure that reduces the noise in product embeddings. The second property relates to the \emph{higher-order relation} and the \emph{functional relation} perspective that is also unique to the e-commerce setting.

We often overlook the problem of false product association in e-commerce. Popular items are likely to co-occur with a large number of irrelevant items, so removing them from the dataset has become a common practice before model training. However, apart from the few globally popular items, false associations are also incurred by various factors including random user behavior, causing the effectiveness of standard data cleaning processes are sensitive to the underlying data distribution. The SGNS embedding algorithm and its variants are vulnerable to misspecified associations because they treat each co-occurrence with equal importance. The false association problem for the embedding model has not been studied before, mainly because it rarely raises concerns in the NLP setting. We explain the domain difference in two folds.
\begin{itemize}
    \item The training data for NLP are extracted from established documents with high reliability. In e-commerce, the training data are often collected from user feedback, so the quality is often uncontrolled and the signal-to-noise ratio is not ideal.
    \item The number of tokenized words in an NLP training corpus if often of several magnitudes smaller than the number of products in an e-commerce dataset, 
    % e.g. tens of thousands versus several million, 
    so their frequency distribution can be different (Figure \ref{fig:illustration}c)\footnote{We use the National Library of Medicine MeSH (Medical Subject Heading) as an example: https://www.dropbox.com/s/sd4yj1uqsqak4n1/d2016.bin?dl=1. }.
    % As a consequence, the most common values in the distribution might not be near the mean, making it difficult to differentiate true association from the false association.  
\end{itemize}

In analogy to the notation of false positive from hypothesis testing, we define false association for the product relatedness measure $R_{i,j} = \log \big\{nN_{i,j}(\Dcal) / \big( N_{i}(\Dcal)N_{j}(\Dcal) \big) \big\}$. 
\begin{definition}
A false association of the product relatedness measure is to observe a large value of $R_{i,j}$ by chance. 
\end{definition}
First notice that the lower bound of $R_{i,j}$ is $-\infty$ so it can take negative values. If product $i$ and $j$ are not related, i.e. $N_{i}(\Dcal)$ is independent of $N_{j}(\Dcal)$, then we can expect $n\Ebb [N_{i,j}(\Dcal)] = \Ebb[N_{i}(\Dcal)] \cdot \Ebb[N_{j}(\Dcal)]$, which implies that $\Ebb[R_{i,j}] = 0$. In theory, we expect the unrelated products to have an ideal zero relatedness measure. However, there are random perturbations when the number of samples is insufficient (which is common in e-commerce dataset). The distributions of $R_{i,j}$ on real-world datasets are provided in Figure \ref{fig:illustration}d, where a proportion of the values are negative. So the question is, how do we characterize our level of confidence when observing a relatively small value of $R_{i,j}$ so that we know it is safe to include the co-occurrence of $(i,j)$ to the training data? 

The asymptotic properties of $R_{i,j}$ can be misleading in this case since the sample size is usually limited. The finite-sample tail bound, which characterizes the deviations from mean, appears to a reasonable choice; however, we need to be cautious because there are caveats in choosing the Hoeffding-type tail bound which tends to be loose on the boundary of $[0,1]$ (see the Appendix for more detail). A careful manipulation using the Chernoff technique leads to a tighter tail bound that can be applied to derive confidence sets.

\begin{lemma}
\label{lemma:CI}
Define $D_{KL}(a \| b) = p\log p/q + (1-p)\log (1-p)/(1-q)$ for $a, b\in (0,1)$. Then for $\forall \epsilon > 0$:
\begin{equation}
\label{eqn:chernoff}
\begin{split}
& \Pbb \big( R_{i,j} \leq  -\epsilon\big) \leq \\ 
& \exp \Big\{ -n D_{KL} \Big(\frac{\Ebb[N_{i}(\Dcal)]\Ebb[N_{j}(\Dcal)]}{n^2e^{-\epsilon}} \, \big\| \,  \frac{\Ebb[N_{i}(\Dcal)] \Ebb[N_{j}(\Dcal)]}{n^2}  \Big)  \Big\}.
\end{split}
\end{equation}
The finite-sample lower confidence bound at $\alpha$-level, i.e. $p\big(R_{i.j} \leq 0 \big) \leq \alpha$, is then given by:
\begin{equation}
\label{eqn:CI}
\begin{split}
    & \log \frac{n^2p_{\alpha}}{\Ebb[N_{i}(\Dcal)]\Ebb[N_{j}(\Dcal)]}, \, \text{where} \\
    & p_{\alpha} = \max \Big\{ p \in [0,1]: \,  D_{KL}\Big( \frac{N_{i,j}(\Dcal)]}{n} \, \Big\| \, p \Big) 
    \leq \frac{\log 1/\alpha}{n} \Big\}.
\end{split}
\end{equation}
\end{lemma}

The proof is mostly technical and we leave it to the appendix. The result in Lemma \ref{lemma:CI} provides a sound criterion for detecting the truly unrelated product pairs. Compare with the other heuristic methods such as removing the top popular products, the confidence-interval approach is theoretically-grounded and is consistent across data distributions. While removing the noisy product pairs reduces false association and improves the quality of embedding, the approach requires the extra step of storing the observed $N_{i,j}(\Dcal)$ and conducting pairwise comparisons, so the computation and memory complexity will depend on the overall sparsity of the data. 

In e-commerce, it is common that complementary products are combined into a combo (bundle) that possesses the overall functionality and contextual meaning of each product. The relations between the bundle and other products are often unchanged by the composition. For instance, \texttt{toothbrush} and \texttt{toothpaste}, are often bundled together, and the \texttt{brush+paste} bundle may inherit their relations with the other personal care products. This type of higher-order relation is essential and unique to e-commerce and should be captured by the product embedding whenever possible. In NLP, for example, when "\texttt{straw}" and "\texttt{berry}" are bundled together, the word "\texttt{strawberry}" has a different contextual meaning. 

We find out that the product relatedness measure $R_{i,j}$ exactly constructs the higher-order relation, so in some sense the product embedding is indeed recognizing and leveraging the higher-order relation to characterizing product relations. Before we present the main discovery, we provide a heuristic definition for the higher-order relation using $R_{i,j}$.
\begin{definition}
Let $I$ be the random variable for products. Given a set of products $\{i_1,\ldots,i_k\}$, if there exists a product $i^* \in \Ical$ such that:
\begin{equation}
\label{eqn:higher-order}
    \Ebb_{I|\{i_1,\ldots,i_k\}}\big[ R_{i^*,I} -  R_{j,I} \big] \geq 0,
\end{equation}
for any other product $j \in \Ical$, then $i^*$ is the higher-order representation of $\{i_1,\ldots,i_k\}$.
\end{definition}
We first provide heuristic understandings for the definition. For example, the \texttt{brush+paste} combo (assuming it exists as a valid product in $\Ical$) could be a higher-order representation of the product set \{\texttt{brush}, \texttt{paste}\}. The expectation term in (\ref{eqn:higher-order}) simply implies that compared with all other choices, the brush+paste combo, on average, has higher relatedness with all the products that may co-occur with \{\text{brush}, \texttt{paste}\} combined:
\begin{equation*}
\sum_{\text{item}}p(\text{item | \{{brush}, {paste}\}})[R_{\text{{brush+paste},item}} - R_{\text{other choice,item}}] > 0.
\end{equation*}
If the definition does lead to an $i^*$ that recovers the relation between $\{i_1,\ldots,i_k\}$ and other items, then the product relatedness measure is indeed capturing the higher-order relation among products.
% \begin{equation}
% \Ebb_{\text{product | \{brush, paste\}}}[R_{\text{brush+paste,product}} - R_{\text{other choice,product}}] > 0.
% \end{equation}
We show that there is an one-to-one mapping between (\ref{eqn:higher-order}) and an optimal information criterion that evaluates the distance of the conditional distributions induced by $i^*$ and $\{i_1,\ldots,i_k\}$.

\begin{claim}
\label{claim:higher-order}
Let $i^*$ be defined by (\ref{eqn:higher-order}) for a meaningful product set $\{i_1,\ldots,i_k\}$ such that $\exists j \in \Ical$: $p\big(j | \{i_1,\ldots,i_k\}\big) > 0$. Then the higher-order product relation can be constructed using the distribution induced by $i^*$ and $\{i_1,\ldots,i_k\}$:
% \begin{equation}
% \label{eqn:higher-order-kl}
% i^* = \arg\min_{i \in \Ical} D_{KL}\Big(p\big(I \,|\, \{i_1,\ldots,i_k\}\big) \,\big\|\, p(I \,|\, i)  \Big),
% \end{equation}
\begin{equation}
\label{eqn:higher-order-kl}
i^* = \arg\min_{i \in \Ical} D_{KL}\Big(p\big(1\big[i \in \Ncal(\{i_1,\ldots,i_k\})\big]\big) \,\big\|\, p\big(1[i \in \Ncal(i^*)]\big)  \Big),
\end{equation}
\end{claim}
The proof is provided in the appendix. We also show that the direction from (\ref{eqn:higher-order-kl}) to (\ref{eqn:higher-order}) also holds true, so there exists a duality between using KL-divergence and product relatedness measure to recognize and construct the higher-order product relations. 

Another interesting property we investigate is inspired by the famous finding from NLP that $\zbf_{\texttt{king}} - \zbf_{\texttt{men}} = \zbf_{\texttt{queen}} - \zbf_{\texttt{women}}$. The simple analogy from NLP is not useful in e-commerce for obvious reasons, however, when a group of functionally-related product pairs are presented, we ask the question of whether their product embeddings can be combined to obtain a meaningful representation for their functional relation. A motivating example will be: (\texttt{TV}, \texttt{remote control}), (\texttt{XBox}, \texttt{handle}) and (\texttt{laptop}, \texttt{mouse}), which all reflect the complementary relation in electronics. So is it possible that $\big(\zbf_{\texttt{TV}} - \zbf_{\texttt{remote control}}\big) + \big(\zbf_{\texttt{XBox}} - \zbf_{\texttt{handle}}\big) + \big(\zbf_{\texttt{laptop}} - \zbf_{\texttt{mouse}}\big)$ captures the functional representation of the complementary relation in electronics?  

We formalize the setup by denoting a relation $r$ by $\stackrel{r}{\rightarrow}$, e.g. handle$\stackrel{\text{complement}}{\rightarrow}$XBox. Again, we focus on the product relatedness measure $R_{i,j}$, where we use the shorthand $\vec{R}_{i} = [R_{i,1}, \cdots, R_{i,\Ical}]$. Following the previous discussion, for a group of product pairs satisfying $\stackrel{r}{\rightarrow}$: $\Dcal_r \equiv \big\{(i,j) | i \stackrel{r}{\rightarrow} j \big\}$, we define $\zbf_r = \sum_{i \stackrel{r}{\rightarrow} j} \vec{R}_{j} - \vec{R}_{i}$. The following claim validates $\zbf_r$ as the representation for relation $r$ according to $\Dcal_r$.
\begin{claim}[Informal]
\label{claim:functional}
Let $\Dcal_r$ and $\zbf_r$ be defined above. For a product pair $i^* \stackrel{r}{\rightarrow} j^*$ not included in $\Dcal_r$, we have:
\begin{equation}
\label{eqn:analogy}
    \vec{R}_{j^*} = \vec{R}_{i^*} + \zbf_r + \mathbf{\epsilon},
\end{equation}
where $\mathbf{\epsilon}$ is the residual term that is negligible under mild conditions.
\end{claim}
The details are relegated to the appendix. Claim \ref{claim:functional} characterizes the generalization property of the functional relation captured by the product relatedness measures. In summary, the product relatedness measure is capable of capturing two of the essential e-commerce-specific relations: the higher-order relation and the functional relation among products, without relying on additional context or supervision. 

Nevertheless, how do the-above explorations on product relatedness measure lead to practical usage other than providing model interpretation? We have shown that the product embeddings are nonlinear projections of the product relatedness measure $R$ so we can not expect (\ref{eqn:higher-order}) and (\ref{eqn:analogy}) to hold for product embedding as well. However, recall that product embedding is also the sufficient dimension reduction for $R$, so $\zbf_i$ is the best compression of $\vec{R}_i$ in the $\Rbb^d$ space. It remains challenging to develop the optimal strategy that best leverages the higher-order relation and functional relation information in the product embedding space. However, we may still proceed with more straightforward (perhaps sub-optimal) practices such as adding the product embeddings in a shopping cart as the cart's representation, and conduct clustering based on the product embedding differences to detect and establish functional relations.

%% file: generalization.tex
Even though we have recognized product embedding as the sufficient dimension reduction of the product relatedness measure, which suggests certain optimally, the performance of product embedding in downstream tasks can still depend on the problem instance, e.g. the embedding dimension $d$. Intuitively speaking, a small $d$ is insufficient for compression, while an overlarge $d$ can introduce extra noise due to the random initialization in the SGNS algorithm. In this paper, we study the product-level downstream tasks, e.g. product classification using the product embedding. Tasks that take a set (sequence) of products as input often employ complicated models, e.g. recurrent neural network, whose generalization performance can be intractable. Notice that the entity-level tasks are also unique to the e-commerce world,  since in NLP, the downstream tasks are for the sentence or document level. 

Before we characterize the generalization performance of product embedding, we point out that our problem is fundamentally different from the ordinary generalization theory in supervised learning:
\begin{itemize}
    \item For the ordinary supervised learning, we study the generalization error from $\Xbf_{\text{train}}$ to $\Xbf_{\text{all}}$, i.e. how the optimal model for the training data performs globally \cite{wainwright2019high}.
    \item In our problem, we study how the model trained with the embedding $\Zbf$ generalizes against the original setting where the the model is trained using $\Xbf$.
\end{itemize}
Specifically, the model using $\Zbf$ is trained by minimizing:
$  
\Lcal(\Zbf) = \Ebb \Big[\frac{1}{|\Ical|} \sum_{i}\phi\big(f_{\theta_1}(\zbf_i), y_i \big)  \Big],
$
and the model using $\Xbf$ is trained by minimizing: 
$
\Lcal(\Xbf) = \Ebb \Big[\frac{1}{|\Ical|} \sum_{i}\phi\big(f_{\theta_2}(\xbf_i), y_i \big)   \Big],
$
where $\phi(\cdot ,\cdot)$ is the loss function that is L-Lipschitz in both arguments\footnote{In the case of binary classification, we do not explicitly assume $y_i$ is categorical. When $y_i$ is given by the score (of the positive class), we simply adapt the multi-class logistic loss, e.g. $\phi\big(f(x_i), y_i\big) = \sigma(y_i)\log\sigma\big(f(x_i)\big) + \big(1-\sigma(y_i)\big)\log\big(1 - \sigma\big(f(x_i)\big)\big)$, where $\sigma(\cdot)$ is the sigmoid function. The loss function is Lipschitz if both $y_i$ and $f(x_i)$ are bounded, which is a mild assumption.}. Here, $\Xbf$ can be given by the product relatedness matrix $[R_{i,j}]_{i,j=1}^{|\Ical|}$, and we use the notation $\Xbf$ to be consistent with the supervised learning literature. It is usually the case that people investigate on the linear model such that $\ybf = \Xbf^{\intercal}\thetabf_2 + \epsilonbf$. However, this generic setting is not applicable to our problem because the data matrix $\Xbf$, e.g. given by $[R_{i,j}]_{i,j=1}^{|\Ical|}$, has a very different geometric structure from product embedding $\Zbf$:
\begin{itemize}
    \item empirical evidence shows that the elements of $\Zbf$ are mostly between $(-1,1)$, while the elements of $\Xbf$ can be unbounded;
    \item the product embedding lies in the Euclidean subspace of $\Rbb^d$, where $\Xbf$ can have arbitrary manifold data structure with a different dimension. 
\end{itemize}
Therefore, we need a standardization protocol that works for both $\Zbf$ and $\Xbf$, which leads us to the singular value decomposition: 
\[ 
\Zbf = \Ubf(Z) \Sigmabf(Z) \Vbf(Z)^{\intercal}, \text{ and } \Xbf = \Ubf(X) \Sigmabf(X) \Vbf(X)^{\intercal},
\]
so $\Ubf(Z)$ and $\Ubf(X)$ are both orthonormal basis. The intuition is that we now think of $\ybf$ as generated by $\Ubf(X)$ to avoid the above-mentioned issues. Then we consider the average-case setting (where the parameters $\theta$ follow an unknown distribution $N(0,\Sigma)$) with a proper scaling $\|\Sigma\|\leq 1$:
\begin{equation}
    \ybf_0 = \Ubf(X)^{\intercal}\thetabf_0, \, \, \ybf = 
    \ybf_0 + \epsilonbf \,\text{ for } \thetabf_0 \sim N(0, \Sigmabf)\, , \epsilonbf \sim N(0, \sigma^2\mathbf{I}).
\end{equation}
As such, we are interested in the average-case loss: \[\Ebb_{\theta_0, \epsilon}[\Lcal(\Zbf)] = \Ebb_{\theta_0, \epsilon}\Big[\sum_i \phi\big(\hat{f}(\zbf_i), \ybf_{0,i} \big)\Big]. \] 
It becomes clear at this point that we study the random-effect setting because the observations are no longer independent, since we assume that they are generated by $\Ubf(X)$ instead of $\Xbf$.
Therefore, in the above average-case loss, the expectation with respect to $\theta_0$ incurs because the "clean testing data" $y_0$ is generated under $\theta_0 \sim N(0, \Sigma)$. As for the $\epsilon$, it also occurs under the expectation because $\hat{f}(\cdot)$ is estimated using the "noisy training data" $\ybf = \ybf_0 + \epsilonbf$ with a given $\ybf_0$, so $\hat{f}(\cdot)$ is a (implicit) function of $\epsilonbf$.
\begin{theorem}
\label{thm:generalization}
Let $f_{\theta_1}(\zbf_i) = \zbf_i^{\intercal}\theta_1$ and $f_{\theta_2}(\xbf_i)=\xbf_i^{\intercal}\theta_2$. The generalization error for product embedding in the above setting follows:
\begin{equation}
\begin{split}
    & \Ebb_{\theta_0,\epsilon}\big[\Lcal(\Zbf) - \Lcal(\Xbf) \big] \leq \\
    & \qquad L \Big\{\Big(tr(\Sigma) - \underline{\lambda(\Sigma)} \big\|\Ubf(X)^{\intercal}\Ubf(Z)\|_{F}^2 \Big)  \big/ |I|  \Big\}^{1/2} + C,
\end{split}
\end{equation}
where $\underline{\lambda(\Sigma)}$ gives the smallest eigenvalue of  $\Sigma$.
\end{theorem}

The proof is relegated to the appendix.
The significance of Theorem \ref{thm:generalization} is that the average-case generalization error of product embedding is controlled by the factor $\big\|\Ubf(X)^{\intercal}\Ubf(Z)\big\|_F^2$, i.e. how well the spectral space of $\Zbf$ aligns with the spectral space of $\Xbf$. While the results reveal the special case under a linear model, it nevertheless provides a novel perspective for understanding the generalization performance of product embedding. We see that the "closeness" between $\Xbf$ and $\Zbf$, which depends on the problem instance (loss function, generating model, learning model, etc.) as we show here, plays a critical part in the generalization bound. We leave it to the future work to derive the results for the more general models.

% Specifically, we see that the generalization of product embedding is a consequence of various factors, e.g. the label generating model, 

% instead of the heuristic idea of checking the approximation error of using $\Zbf$ (and $\tilde{\Zbf}$) to reconstruct $\Xbf$, we may as well see how close their spectral spaces align. 

%% file: experiment.tex
We design our experiments to provide empirical supports for our theoretical results, as well as to shed insights for future research and application with product embedding. All the reported numerical results are computed from ten repetitions. We use $d=32$ unless specified otherwise. 

\textbf{Dataset.} 

We use the public \textsl{Instacart} dataset\footnote{https://www.instacart.com/datasets/grocery-shopping-2017} for reproducibility. As for the experiments where the resource and information in public dataset do not satisfy our need, we use the proprietary \textsl{Walmart.com} datasets which we have full access. The \textsl{Instacart} data consists of $\sim$50,000 grocery products, with the shopping records of $\sim$200 thousand users and 3 million orders. The product catalog information, i.e. the \emph{category} and \emph{department} tags, can be used as labels for downstream classification task. We experiment on two types of data-generating mechanism $\Dcal(\Ncal)$: 
\begin{itemize}
    \item \textbf{Sequences}. Choosing the users' sequential impression (purchase) as input data structure, and the neighborhood is defined by using the five previous purchases;
    \item \textbf{Graphs}. We build the \emph{undirected} weighted graph using session-based purchase data. The neighborhood is then obtained by sampling five nodes according to the random walk outcome like the \emph{Node2vec} \cite{grover2016node2vec}.
\end{itemize}

\textbf{Optimality of sufficient dimension reduction.} 

Directly examining the optimally of product embedding, which is the sufficient dimension reduction as we show in Claim \ref{claim:sdr}, is infeasible so we need to rely on certain tasks. We consider:\\
\textbf{Task 1}. The next-item recommendation, where we use all but the last item for training, the second-to-last item for validation (if needed), and the last item for testing. The inner product\\ $\langle \zbf_{\text{last item}}, \zbf_{\text{next item}} \rangle$ is used to rank the candidate items; \\
\textbf{Task 2}. The items' \emph{department} classification, where we simply employ the multi-class logistic regression: $\text{softmax}\big(\zbf_i^{\intercal}\Theta\big)$, as the classifier. The model is trained using the \textsl{Scikit-Learn} package.

We show that product embedding achieves better performance than using the least-square linear dimension reduction (\ref{eqn:grad-ls}) of the same $R_{i,j}$, when the embedding dimension is also $d=32$. When estimating $R_{i,j}$ from the data $\Dcal(\Ncal)$, we treat $\hat{r}_{i,j}$ as missing value if $(i,j)$ never co-occurs, and we treat the negative estimated values as zero (because in theory, $R_{i,j}=0$ if the two products are unrelated). The comparisons of the results are provided in Table \ref{tab:main}. We see that product embedding outperforms the linear dimension reduction in both tasks, supporting the optimality of product embedding as sufficient dimension reduction.
\begin{table}[hbt]
\small
    \centering
    \begin{tabular}{c|c|c|c|c}
    \hline
        Data & \multicolumn{2}{c}{Instacart} & \multicolumn{2}{c}{Walmart.com}  \\ \hline \hline
         & \multicolumn{4}{c}{\textbf{recommendation}} \\ \hline \hline
        Metric & AUC & NDCG & Recall@10 & NDCG@10 \\ \hline
        LDR(seq) & 0.939(.008) & 0.142(.004) & 0.112(.005) & 0.058(.002)\\ 
        \textbf{Emb}(seq) & \emph{0.954}(.005) & \emph{0.160}(.004) & \emph{0.155}(.004) & \emph{0.079}(.002) \\
        LDR(graph) & 0.929(.010) &0.139(.006) & 0.107(.008) & 0.053(.004) \\
        \textbf{Emb}(graph) & \emph{0.948}(.008) &\emph{0.155}(.005)  &\emph{0.131}(.006) & \emph{0.070}(.004) \\ \hline \hline
         & \multicolumn{4}{c}{\textbf{classification}} \\ \hline \hline
        Metric & micro-F1 & macro-F1 & micro-F1 & macro-F1 \\ \hline
        LDR(seq) & 0.483(.010) & 0.342(.013) & 0.312(.017) & 0.253(.013) \\ 
        \textbf{Emb}(seq) & \emph{0.509}(.008) & \emph{0.470}(.010) & \emph{0.395}(.018) & \emph{0.313}(.011) \\
        LDR(graph) & 0.487(.012) & 0.345(.019) & 0.316(.022) & 0.260(.017) \\
        \textbf{Emb}(graph) & \emph{0.513}(.010) & \emph{0.476}(.014) & \emph{0.404}(.018) & \emph{0.317}(.014) \\ \hline \hline
    \end{tabular}
    \caption{\small Recommendation and classification results. \textsl{LDM} and \textsl{Emb} denote using linear dimension reduction and SGNS algorithm, and \textsl{seq} and \textsl{graph} indicates the data structure.}
    \label{tab:main}
\end{table}

\textbf{Improvement of removing false associations.} 

\begin{figure}
    \centering
    \includegraphics[width=0.75\linewidth]{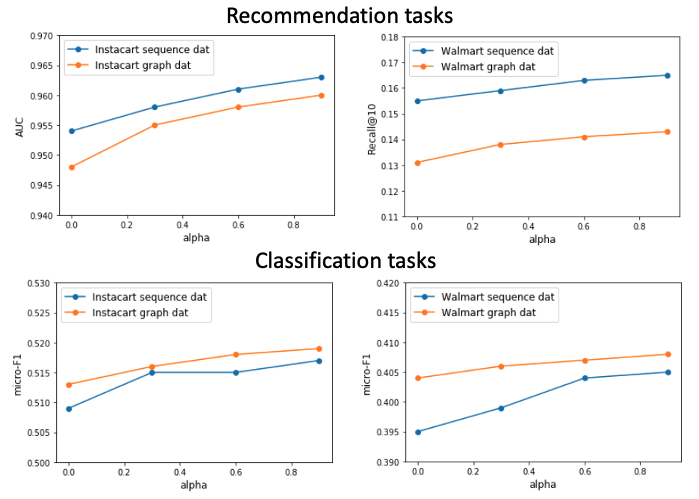}
    \caption{\small The effectiveness of removing false associations according to the confidence bound in Lemma \ref{lemma:CI}, for both the recommendation and classification task.}
    \label{fig:false-association}
\end{figure}

According to the finite-sample confidence interval from Lemma \ref{lemma:CI}, we do a full scan of the training data and remove the suspicious product co-occurrences with confidence level $\alpha \in \{0.3, 0.6, 0.9\}$ by treating them as zero. We then train the product embedding (using SGNS) and examine the performance via the same two tasks. The results are shown in Figure \ref{fig:false-association}, where we see that a higher confidence level does lead to improvements for both tasks. The degree of improvement is more significant by moving from small to medium $\alpha$. The improvement curves gets flattened under large $\alpha$, which can be caused by removing too much training data. The proposed preprocessing approach is overall effective, but the tradeoff between the quality and the size of data should be judged case-by-case.

\textbf{Cart-page recommendation: an example on exploring the higher-order relation.} 

We obtain $\sim$100,000 shopping cart snapshots with the customers' continual shopping records from \textsl{Walmart.com}. To examine the heuristic that by adding individual product embedding (\textbf{Add}) as the cart's embedding for the next-item (provided by the continual-shopping record) recommendation, we compare with: \\
\textbf{Baseline1}. Randomly select an item from the cart and use its embedding as the cart's embedding; \\
\textbf{Baseline2}. Use the most-recent added item's embedding as the cart's embedding; \\
\textbf{Oracle}. The best item embedding in retrospect, i.e. after we observe the user's next move and select the best item in the cart, as an ad-hoc approach; \\
\textbf{Enhanced}. Apply a simple dot-product self-attention to obtain the weights and use the weighted sum of the item embedding.

The cart-page recommendation performance is provided in Table \ref{tab:cart}. Compared with using a single product embedding, even when the single product is given by the oracle, combining the product embeddings leads to better performance under the simple addition. When the weight for each product is more carefully chosen, such as by using the dot-product attention mechanism, we observe a further improvement. The result is not surprising, since combining individual product embedding is becoming common in personalized recommendation. In this paper, we further justify this approach via the lens of higher-order product relations.

\begin{table}[htb]
\small
    \centering
    \begin{tabular}{p{0.8cm}|c|c|c|c|c}
    \hline
         & Baseline1 & Baseline2 & Oracle & Add & Enhanced  \\ \hline \hline
        Recall @10 & .051(.011) & .064(.003) & .093(.002) & .101(.002) & .112(.003) \\
        NDCG @10 & .027(.004) & .030(.001) & .038(.001) & .044(.001) & .046(.001) \\ \hline 
    \end{tabular}
    
    \vspace{0.1cm}
    
    \begin{tabular}{p{1.6cm}|c|c|c|c}
    \hline
        dimension & d=8 & d=16 & d=32 & d=64 \\ \hline \hline
        $S(Z, X)$ & 0.043 & 0.105 & 0.144 & 0.212 \\
        micro-F1 & 0.206 & 0.358  & 0.404  & 0.439  \\ \hline
    \end{tabular}
    \caption{\small \textbf{Upper}: the cart-page recommendation performances; \textbf{Lower}: the experiments for the generalization of embedding in classification task. Both experiments are conducted on the \emph{Walmart.com} data.}
    \label{tab:cart}
\end{table}

\textbf{Detecting product functional relations.} 

\begin{figure}
    \centering
    \includegraphics[width=\linewidth]{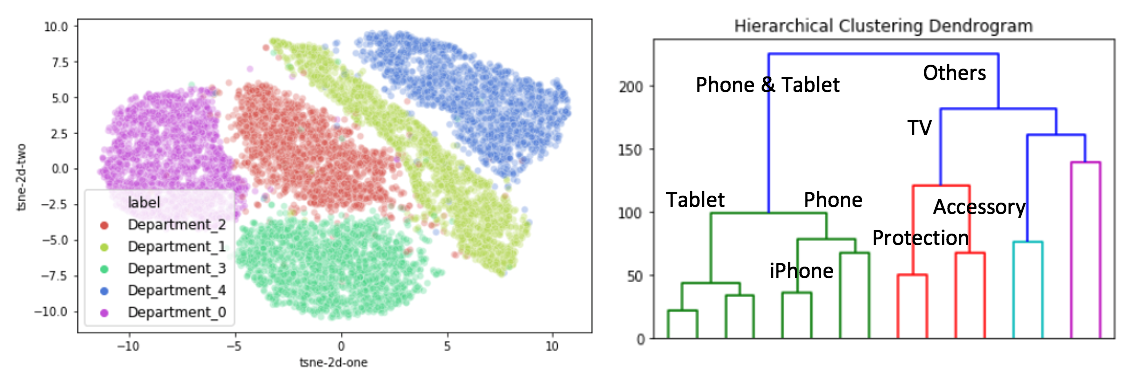}
    \caption{\small \textbf{Left}: the clustering results (visualized after a t-sne projection to 2D) that detects the different product functional relations under each product department; \textbf{Right}: hierarchical clustering result (the branches are labelled after cross-checking with the ground-truth item catalog).}
    \label{fig:clustering}
\end{figure}

Here, we provide a brief demonstration on some interesting results we obtained by clustering the embedding difference of product pairs, i.e. $\zbf_{\text{anchor}} - \zbf_{\text{reco}}$, for the 1,000 most popular anchor items with their top-10 recommendation (obtained by using the inner products of product embedding). All the items are selected from the \textsl{electronics} catalog on \textsl{Walmart.com}. We conduct both the K-means clustering and hierarchical clustering, with results shown in Figure \ref{fig:clustering}. Under the correctly-specified number of clusters, K-means exactly detects the different functional relations for each department of electronic products. To make sense of the hierarchical clustering result, we do a manual cross-checking to label the different branches from the dendrogram, leveraging the true department tag for the items. We find that hierarchical clustering keeps showing refined detection of finer-granulated product functional relations. Our discovery supports the result in Claim \ref{claim:functional}, and provide insights for understanding product relations via the pairwise embedding differences.

\textbf{Generalization performance of downstream tasks.} 

To support our generalization results, we vary the embedding dimension $d \in \{8, 16, 32, 64\}$ as a control factor to obtain product embeddings that give different spectral \emph{alignment score} $S(Z, X):=\big\|\Ubf(X)^{\intercal}\Ubf(Z)\big\|_F^2$, where the data matrix $X$ is given by the estimated $[R_{i,j}]$ matrix. Here, we specifically study the item classification task. As we conjectured, a larger dimension does lead to a higher score within the range we consider, where the results are provided in Table \ref{tab:cart}. We see that a higher spectral alignment score leads to better downstream classification performance, where the classifier is logistic-regression so the empirical result is consistent with our theoretical justifications. Our discussion may lead to methods that practically chooses $d$ in a data-adaptive fashion (which is out of the scope of this paper).

% {\large{\textbf{Conclusion}}}. 
\section{Conclusion}
We thoroughly study the theoretical backgrounds of product embeddings by answering what they are, how they are unique to e-commerce, and why they are useful in downstream tasks. With both the technical derivations and intuitive explanations, we hope this paper provides tools and reference for model interpretation and understanding, as well as developing more advanced techniques for representation learning in e-commerce.

%% file: proofs.tex
We provide the proofs in this part of the paper.

\subsection{Proof for Claim \ref{claim:sdr}}.
\begin{proof}
We start by considering the co-occurrence random variable $O$ as following a Bernoulli distribution such that $p(O=1) = \beta$, where $\beta \in (0,1)$ characterizes the global probability of having a positive sample, which for the SGNS it is given by $1/k+1$, since for each positive sample we generate $k$ negative samples. 

Consequently, we have $p(i,j | O) = \left\{ \begin{array}{lll}
    p_{i}(\Dcal)p_{j}(\Dcal) & if & O=0,  \\
    p_{i,j}(\Dcal) & if & O=1 .
\end{array} \right.$

Applying the Bayes rule, we immediately have:
\begin{equation}
\label{eqn:redef-cooccur}
p(O=1 \,|\, i,j) = \frac{\beta p_{i,j}(\Dcal)}{\beta p_{i,j}(\Dcal) + (1-\beta)p_{i}(\Dcal)p_{j}(\Dcal)}.
\end{equation}

We point out that this definition of co-occurrence probability does not contradict the original definitions in (\ref{eqn:setup}), since we have:
\[
\frac{\beta p_{i,j}(\Dcal)}{\beta p_{i,j}(\Dcal) + (1-\beta)p_{i}(\Dcal)p_{j}(\Dcal)} = \frac{1}{1+ \exp\Big( -\log \big( \frac{\beta p_{i,j}(\Dcal)}{(1-\beta)p_{i}(\Dcal)p_{j}(\Dcal)} \big) \Big)} = \sigma(\tilde{R}_{i,j}),
\]
where $\tilde{R}_{i,j} = R_{i,j} + \log\frac{\beta}{1-\beta}$ is a shifted version of the original product relatedness measure. The impact of using the shifted original product relatedness on the loss function is negligible, because we now have:
\[
\ell(\Dcal) = \sum_{i, j \in \Ical} N_{i,j}(\Dcal) \log \sigma(\zbf_i^{\intercal}\tilde{\zbf}_j) + \frac{k}{n} N_i(\Dcal) N_j(\Dcal) \log \sigma(-\zbf_i^{\intercal}\tilde{\zbf}_j) + f(\beta(k)),
\]
and $f(\beta(k)) := f(k/(1+k))$ is irrelevant to the optimization variables.

The first-order necessary condition for the global optimal of the SGNS objective without dimension constraint, which we denote by:
\begin{equation*}
\Zbf^*, \tilde{\Zbf^*} = \arg\min_{\Zbf, \tilde{\Zbf}}\ell(\Dcal),
\end{equation*}
is given by $\nabla_{\zbf_i} \ell(\Dcal) = 0$ and $\nabla_{\tilde{\zbf}_i} \ell(\Dcal) = 0$, for $i=1,\ldots,\|\Ical\|$.
According to (\ref{eqn:grad-sgns}), the first-order condition implies that $\langle \zbf_i^*, \tilde{\zbf}_i^* \rangle = R_{i,j}$ for all $i,j \in \Ical$, because $\sigma(\cdot)$ is a strictly increasing function. Hence, $\ell(\Dcal; \Zbf^*, \tilde{\Zbf^*}) := \min_{\Zbf, \tilde{\Zbf}}\ell(\Dcal)$ is a fixed quantity that only depends on $\Dcal$. Therefore, minimizing $\ell(\Dcal)$ is equivalent to finding :
\[
\arg\min_{\Zbf, \tilde{\Zbf} \in \Rbb^d} \big\{\ell(\Dcal) -  \min_{\Zbf, \tilde{\Zbf}}\ell(\Dcal) \big\} = \arg\min_{\Zbf, \tilde{\Zbf} \in \Rbb^d} \big\{\ell(\Dcal) -  \ell(\Dcal; \Zbf^*, \tilde{\Zbf^*}) \big\}.
\]

According to the above argument on the implication of the first-order condition, the term $\ell(\Dcal; \Zbf^*, \tilde{\Zbf^*})$ is given by:
\[
\sum_{i, j \in \Ical} N_{i,j}(\Dcal) \log \sigma(R_{i,j}) + \frac{k}{n} N_i(\Dcal) N_j(\Dcal) \log \sigma(-R_{i,j}).
\]

Recall that $q\big(O \, \big| \, \Dcal;\, \Zbf,\tilde{\Zbf}\big)$ is the co-occurrence probability computed by the embedding as in (\ref{eqn:setup}), and $p\big(O \, \big| \, \Dcal;\, R\big)$ is the co-occurrence probability when the embedding matrices are given by the product relatedness matrix (that happens for the unconstrained global optimum). By rearranging terms and extracting the factor of $n(k+1)$ to the front, it holds that:
\begin{equation}
\begin{split}
    & n(k+1)\big(\ell(\Dcal) -  \min_{\Zbf, \tilde{\Zbf}}\ell(\Dcal) \big) \\ 
    & \propto \sum_{i,j \in \Ical} \Big\{\frac{1}{k}p_{i,j}(\Dcal)\log\frac{\sigma(R_{i,j})}{\sigma(\zbf_i^{\intercal}\zbf_j)} +  (1-\frac{1}{k})p_i(\Dcal)p_j(\Dcal)\log\frac{\sigma(-R_{i,j})}{\sigma(-\zbf_i^{\intercal}\zbf_j)} \Big\} \\
    & = \sum_{i,j \in \Ical} \Big\{\frac{1}{k}p_{i,j}(\Dcal)\log\frac{p\big(O_{i,j}=1\, | \,R\big)}{q\big(O_{i,j}=1\, | \,\zbf_i,\tilde{\zbf}_j\big)} +  (1-\frac{1}{k})p_i(\Dcal)p_j(\Dcal)\log\frac{p\big(O_{i,j}=0\, | \,R\big)}{q\big(O_{i,j}=0\, | \,\zbf_i,\tilde{\zbf}_j\big)} \Big\} \\
    & \overset{(a)}{=} \sum_{\substack{\alpha \in \{0,1\} \\ (i,j)\in \Dcal}} \Big\{ p\big(O_{i,j}=\alpha\big) \log\frac{p\big(O_{i,j}=\alpha\, | \,R\big)}{q\big(O_{i,j}=\alpha\, | \,\zbf_i,\tilde{\zbf}_j\big)} \Big\} \\
    & = D_{KL}\Big( q\big(O \, \big| \, \Dcal;\, \Zbf,\tilde{\Zbf}\big) \, \big\|\, p\big(O \, \big| \, \Dcal;\, R\big) \Big),
\end{split}
\end{equation}
where we use the shorthand $O_{i,j}=\alpha$ to denote the event $O=\alpha, i,j $; and step $(a)$ follows from (\ref{eqn:redef-cooccur}). 

Therefore, solving for $\min_{\Zbf, \tilde{\Zbf} \in \Rbb^d}\ell(\Dcal)$ is equivalent to finding:
\[
\arg\min_{\Zbf, \tilde{\Zbf} \in \Rbb^d}D_{KL}\Big( q\big(O \, \big| \, \Dcal;\, \Zbf,\tilde{\Zbf}\big) \, \big\|\, p\big(O \, \big| \, \Dcal;\, R\big) \Big),
\]
which concludes the proof.
\end{proof}

\subsection{Proof for Lemma \ref{lemma:CI}}
\begin{proof}

We first prove the auxiliary case that with $X_1,\ldots,X_n$ being a sequence of independent Bernoulli random variables under mean $\mu$, and $\hat{\mu}$ given by: $\frac{1}{n}\sum_{i=1}^n X_i$, it holds for any $\epsilon \in [0,1-\mu]$ that:
\[
\Pbb \big(\hat{\mu} \leq \mu - \epsilon\big) \leq \exp\big(-n D_{KL}(\mu-\epsilon \,\|\, \mu)\big). 
\]

The above result is straightforward by using the Cramer-Chernoff bounding technique:
\begin{equation*}
\begin{split}
    \Pbb \big(\hat{\mu} & \leq \mu - \epsilon\big) \leq \Pbb \Big(\exp\Big(\lambda \sum_{i=1}^n (\mu - X_t) \Big) \geq \exp(\lambda n \epsilon) \Big) \\ 
    & \leq \frac{\Ebb \big[\exp\Big(\lambda \sum_{i=1}^n (\mu - X_t) \big]}{\exp(\lambda n \epsilon)} \\ 
    & = \big\{\mu \exp(\lambda(1-\mu -\epsilon)) + (1-\mu) \exp(\lambda(\mu+\epsilon)) \big\}^n,
\end{split}
\end{equation*}
holds for any $\lambda>0$. Note that the above objective is convex in $\lambda$ (when $\lambda>0$), so the unique minimizer is given by: $\lambda^* = \log\frac{(\mu+\epsilon)(1-\mu)}{\mu(1-\mu-\epsilon)}$. We plug $\lambda^*$ back to the above expression and obtain:
$\Pbb \big(\hat{\mu} \leq \mu - \epsilon\big) \leq \exp\big(-n D_{KL}(\mu-\epsilon \,\|\, \mu)\big).$ 

We now let $\mu_{i,j} = \frac{\Ebb N_i(\Dcal)N_j(\Dcal)}{n^2}$ and $\hat{\mu} = \frac{N_i(\Dcal)N_j(\Dcal)}{n}$, which are the mean and empirical average of the Bernoulli random variables defined in our setting. It then holds that:
\begin{equation*}
\begin{split}
    & \Pbb\big(\hat{\mu}_{i,j} \leq \mu_{i,j} - \epsilon \big) \leq \exp\Big(-nD_{KL}\Big(\frac{\Ebb N_i(\Dcal)N_j(\Dcal)}{n^2} - \epsilon \, \big\|\, \frac{\Ebb N_i(\Dcal)N_j(\Dcal)}{n^2} \Big) \Big) \\
    & \Leftrightarrow \Pbb\big(\log\frac{\hat{\mu}_{i,j}}{\mu_{i,j}} \leq \log\big(1-\frac{\epsilon}{\mu_{i,j}} \big) \big) \leq \exp\big(-n D_{KL}(\mu_{i,j}-\epsilon \,\big\|\, \mu_{i,j}) \big), \quad \text{define } \tilde{\epsilon}: \epsilon=\mu_{i,j}(1+\exp(\tilde{\epsilon})) \\
    & \Leftrightarrow \Pbb\big(\log\frac{\hat{\mu}_{i,j}}{\mu_{i,j}} \leq \tilde{\epsilon} \big) \leq \exp\Big(-nD_{KL}\Big(\frac{\mu_{i,j}}{\exp(-\tilde{\epsilon})} \, \big\| \, \mu_{i,j} \Big) \Big).
\end{split}
\end{equation*}

Then notice that $D_{KL}(\cdot, \, \| \, \mu)$ is decreasing on $[0,\mu]$, so if $\epsilon$ is the (unique) solution of $D_{KL}(\mu - \epsilon \, \| \, \mu) = \alpha$ on $[0,\mu]$ for some $\alpha \in [0, D_{KL}(0 \, \| \, \mu)]$, it holds that:
\[ 
\big\{D_{KL}(\hat{\mu} \, \| \, \mu) \geq \alpha, \hat{\mu} \leq \mu  \big\} = \{\hat{\mu} \leq \mu - \epsilon, \hat{\mu} \leq \mu\} = \{\hat{\mu} \leq \mu - \epsilon\}.
\]

Consequently, using the result from the beginning of the proof, we have:
\[
\Pbb \big( D_{KL}(\hat{\mu} \, \| \, \mu) \geq \alpha, \hat{\mu} \leq \mu  \big) \leq \exp(-n\alpha).
\]

Next, we take $\tilde{p}_{\alpha} = \max\{\mu\in[0,1]: D_{KL}(\hat{\mu}\,\|\,\mu) \leq \alpha\}$, and it is straightforward to verify that: $\tilde{p}_{\alpha} \geq \hat{\mu}$ and $D_{KL}(\hat{\mu}\,\|\,\cdot)$ is strictly increasing on $[\hat{\mu}, 1]$. Therefore, it holds that:
\[
\{\mu\geq\tilde{p}_{\alpha} \} = \{\mu\geq\tilde{p}_{\alpha}, \mu \geq \hat{\mu}\} = \{D_{KL}(\hat{\mu}\,\|\,\mu)\geq D_{KL}(\hat{\mu}\,\|\,\tilde{p}_{\alpha}),\mu \geq \hat{\mu}\} = \{D_{KL}(\hat{\mu}\,\|\,\mu)\geq \alpha,\mu \geq \hat{\mu}\},
\]
which directly leads to: $\Pbb(\mu \geq \tilde{p}_{\alpha}) \leq \exp(-n\alpha)$. 

Now we replace $\mu$ by $\Ebb N_{i}(\Dcal) \Ebb N_{i}(\Dcal)/n^2$ and define $\tilde{\alpha} = \exp(-n\alpha)$, which gives the desired result in (\ref{eqn:CI}).

\end{proof}

\subsection{Proof for Claim \ref{claim:higher-order}}
\label{sec:claim2}
\begin{proof}
Recall from Definition 2 that $I$ is the random variable for products. Given a set of products $\{i_1,\ldots,i_k\}$, if there exists a product $i^* \in \Ical$ such that:
\begin{equation*}
\label{eqn:higher-order}
    \Ebb_{I|\{i_1,\ldots,i_k\}}\big[ R_{i^*,I} -  R_{j,I} \big] \geq 0,
\end{equation*}
for any other product $j \in \Ical$, then $i^*$ is the higher-order representation of $\{i_1,\ldots,i_k\}$.

Define shorthand $\vec{I} = \{i_1,\ldots,i_k\}$ as the combo of the $k$ items, and $p(i|j) := p\big(\ind [j \in \Ncal(i)])$. Similarly, we use the shorthand: $p(\vec{I}):= p(i_1,\ldots,i_k)$ and $p(i|j) = p(\ind[i\in\Ncal(j)])$. 

By rearranging terms and apply basic algebraic manipulations, it holds that:

\begin{equation*}
\begin{split}
D_{KL}\Big(p\big(\ind\big[i \in \Ncal(\{i_1,\ldots,i_k\})\big]\big) \,\big\|\, p\big(\ind[i \in \Ncal(i^*)]\big)  \Big) & = \sum_{e\in\Ical} p(e|\vec{I}) \log \frac{p(e|\vec{I})}{p(e | i^*)} \\
& = \sum_{e\in\Ical}p(e|\vec{I}) \Big( \log \frac{p(\vec{I})}{\prod_{i \in \vec{I}}p(i)} - \log \frac{p(\vec{I}|e)}{\prod_{i\in \vec{I}}p(i|e)}  + \log \frac{p(i^* | e)}{p(i^*)} - \log \prod_{i\in \vec{I}}\frac{p(i|e)}{p(i)} \Big), \\
    &= \sum_{e\in\Ical}p(e|\vec{I}) \Big( \log \frac{p(\vec{I})}{\prod_{j \in \vec{I}}p(j)} - \log \frac{p(\vec{I}|e)}{\prod_{j\in \xbf}p(j|e)} + \Rbf_{i^*,e} - \sum_{i \in \vec{I}}\Rbf_{i,e} \Big) \\
    & = \Ebb_{e|\vec{I}}\Big[ \log \frac{p(\vec{I})}{\prod_{j \in \vec{I}}p(j)} - \log \frac{p(\vec{I}|e)}{\prod_{j\in \vec{I}}p(j|e)} - \sum_{i \in \vec{I}}\Rbf_{i,e} \Big] + \Ebb_{e | \vec{I}} \big[\Rbf_{i^*,e} \big].
\end{split}
\end{equation*}

Notice that the first term in the above expression is independent of $i^*$, and as a consequence, when:
\[ 
D_{KL} \big(p(I | \{i_1,\ldots,i_k\}) \,\big\|\, p(I | i^*)\big) \leq D_{KL} \big(p(j | \{i_1,\ldots,i_k\}) \,\big\|\, p(I | j)\big)
\]
it must hold that: $\Ebb_{I | \{i_1,\ldots,i_k\}} \big[\Rbf_{i^*,I} - \Rbf_{j,I}  \big] \geq 0, \forall j \in \Ical$, which exactly recovers (\ref{eqn:higher-order}). 

It is easy to see that each step in the above derivation is invertible, so we can also obtain the statement in the claim by starting from Definition 2. Together, they give the desired results.
\end{proof}

\subsection{Proof for Claim \ref{claim:functional}}

\begin{proof}
We follow the setup from \ref{sec:claim2}, and define the following shorthand:
\[
\tau(\vec{I}) = \log\frac{p(\vec{I})}{\prod_{i \in \vec{I}} p(i)} \text{ and } \tau(\vec{I} | e) = \log \frac{p(\vec{I}|e)}{\prod_{i \in \vec{I}} p(i|e)}.
\]
Notice that $\tau(\vec{I})$ is measuing the mutual independency among $\vec{I}$, and $ \tau(\vec{I} | e)$ measure the conditional independence of $\vec{I}|e$ for $e \in \Ical$. The mutual independency and conditional independence terms are usually very small in the e-commerce setting, and the degree to which this assumption is valid decides the quality of the approximation in the statement.

Recall that for a group of product pairs satisfying $\stackrel{r}{\rightarrow}$: $\Dcal_r \equiv \big\{(i,j) | i \stackrel{r}{\rightarrow} j \big\}$, we define $\zbf_r = \sum_{i \stackrel{r}{\rightarrow} j} \vec{R}_{j} - \vec{R}_{i}$. We further decompose $\Dcal_r$ into $\Dcal_r^{(+)} \cup \Dcal_r^{(-)}$, where $\Dcal_r^{(+)} = \{i \,|\, \exists j\in\Ical \text{ s.t. } (i,j) \in \Dcal_r\}$ and $\Dcal_r^{(i)} = \{j \,|\, \exists i\in\Ical \text{ s.t. } (i,j) \in \Dcal_r\}$. We make the decomposition because the product functional relations are often asymmetric, which means $i^* \overset{r}{\to} j^* \not\Leftrightarrow j^* \overset{r}{\to} i^*$.

It holds for any $e \in \Ical$ that:
\begin{equation}
\label{eqn:appendix1}
\begin{split}
   & \Rbf_{i^*,e} - \Rbf_{j^*,e} \\
   & = \log \frac{p(e|i^*)}{p(e|j^*)} + \log \prod_{q^+ \in \Dcal_r^{(+)}}\frac{p(e|q^+)}{p(e|q^+)} + \log \prod_{q^- \in \Dcal_r^{(-)}}\frac{p(e|q^-)}{p(e|q^-)} \\ 
   & = \sum_{q^+ \in \Dcal_r^{(+)}}\log p(q^+|e) -  \sum_{q^- \in \Dcal_r^{(-)}}\log p(q^-|e)  + \log \frac{\prod_{q^- \in \Dcal_r^{(-)}\cup i^*} p(e|q^-)}{\prod_{q^+ \in \Dcal_r^{(+)}\cup j^*} p(e|q^+)} \\ 
   & = \sum_{q^+ \in \Dcal_r^{(+)}}\Rbf_{q^+,i^*} - \sum_{q^- \in \Dcal_r^{(-)}}\Rbf_{q^-,j^*} + \underbrace{
   \log\frac{p\big(e|i^*,\Dcal_r^{(-)}\big)}{p\big(e|j^*,\Dcal_r^{(+)}\big)} - \tau\big(i^*\cup \Dcal_r^{(-)}\,\big|\, e\big) + \tau\big(j^*\cup \Dcal_r^{(+)}\,\big|\, e\big) - \tau\big(i^*\cup \Dcal_r^{(-)}\big) + \tau\big(j^*\cup \Dcal_r^{(+)}\big)}_{\epsilon}.
\end{split}
\end{equation}
Consequently, we reach:
\[
\vec{R}_{j^*} = \vec{R}_{i^*} + \zbf_r + \mathbf{\epsilon},
\]
where $\zbf_r = \sum_{i \stackrel{r}{\rightarrow} j} \vec{R}_{j} - \vec{R}_{i}$ and the $\epsilon$ term is highlighted in the above expression.

As we mentioned in the beginning, in the expression of $\epsilon$, the mutual independence and conditional independence terms are usually negligible compared with $\zbf_r$, so the approximation of $\vec{R}_{j^*} \approx \vec{R}_{i^*} + \zbf_r$ can hold with fine granularity under a large sample size. 
\end{proof}

\subsection{Proof for Theorem \ref{thm:generalization}}

\begin{proof}
Recall that the setting we study is:
\[
\ybf = \Ubf(\Xbf)\thetabf + \epsilonbf, \quad \thetabf \sim N(0, \Sigmabf), \quad \epsilonbf \sim N(0, \sigma^2\mathcal{I}),
\]
and we consider the loss function such as $\phi(y, \hat{y}) = \sigma(y)\log\sigma(y) + (1-\sigma(y))\log(1-\sigma(y))$, and assume that the loss function is $L$-Lipschitz in both arguments. 

Let the noise-free label be given by: $\ybf_0 = \Ubf(\Xbf)\thetabf$. We define: $\thetabf^*_{1} = \arg\min \sum_{i=1}^{|\Ical|} \phi (\Xbf_i\thetabf, y_i)$, $\thetabf^*_{2} = \arg\min \sum_{i=1}^{|\Ical|} \phi (\Zbf_i\thetabf, y_i)$ to be the optimum when using $\Xbf$ and $\Zbf$ as the features, and their predictions are given by: $\ybf_1 = \Xbf\thetabf^*_1$ and $\ybf_2 = \Zbf\thetabf^*_2$. Therefore, the empirical training loss using $\Xbf$ is given by: $L(\ybf_1,\ybf) = \frac{1}{|\Ical|}\sum_{i=1}^{|\Ical|}\phi(\ybf_{1,i}, \ybf_i)$, and the empirical training loss using $\Zbf$ is given by: $L(\ybf_2,\ybf) = \frac{1}{|\Ical|}\sum_{i=1}^{|\Ical|}\phi(\ybf_{2,i}, \ybf_i)$. It is easy to verify that $L(\cdot,\cdot)$ is $L/\sqrt{|\Ical|}$-Lipschitz in both arguments.

On the other hand, recall that the average risk $\Lcal$ using $\Xbf$ and $\Zbf$ for predicting the "clean label" is given by: $\Lcal(\Xbf) = \Ebb_{\epsilonbf}\Big[ \frac{1}{|\Ical|}\sum_{i=1}^{|\Ical|} \phi(\Xbf_i\thetabf_1^*, \ybf_i) \Big]$ and $\Lcal(\Zbf) = \Ebb_{\epsilonbf}\Big[ \frac{1}{|\Ical|}\sum_{i=1}^{|\Ical|} \phi(\Zbf_i\thetabf_2^*, \ybf_i) \Big]$. 

By the definition, it is easy to verify that:
\begin{equation}
\label{eqn:generalization-eqn1}
    \Lcal(\Xbf) = \Ebb_{\epsilonbf}\Big[ \frac{1}{|\Ical|}\sum_{i=1}^{|\Ical|} \phi(\Xbf_i\thetabf_1^*, \ybf_i) \Big] \geq \Ebb_{\epsilonbf}\big[L(\ybf_0, \ybf_0) \big] = L(\ybf_0, \ybf_0).
\end{equation}

Then it holds for all $\thetabf_2$ that:
\begin{equation*}
\label{eqn:generalization-eqn2}
\begin{split}
    \Lcal(\Zbf) & = \Ebb_{\epsilonbf}\big[L(\Zbf\thetabf_2^*, \ybf_0) \big] \\
    & \leq \Ebb_{\epsilonbf}\Big[L(\Zbf\thetabf_2^*, \ybf) + \frac{L}{\sqrt{|\Ical|}}\|\epsilonbf\|_2 \Big]\quad (\text{using the Lipschitz condition of }L) \\
    & \leq  \Ebb_{\epsilonbf}\Big[L(\Zbf\thetabf_2, \ybf) + \frac{L}{\sqrt{|\Ical|}}\|\epsilonbf\|_2 \Big]\quad (\text{by the definition of }\thetabf_2^*) \\
    & \leq \Ebb_{\epsilonbf}\Big[L(\Zbf\thetabf_2, \ybf_0) \Big] + 2 \Ebb_{\epsilonbf}\Big[\frac{L}{\sqrt{|\Ical|}}\|\epsilonbf\|_2 \Big] \quad (\text{again by using the the Lipschitz condition of }L) \\ 
    & \leq L(\Zbf\thetabf_2, \ybf_0) + 2L\sigma \quad (\text{by the definition of }\epsilonbf).
\end{split}
\end{equation*}

Also, by the textbook derivation, it holds that:
\begin{equation}
\label{eqn:generalization-eqn3}
\min_{\thetabf_2}\|\Zbf\thetabf_2 - \ybf_0\|_2^2 = \big\|\Zbf(\Zbf^{\intercal}\Zbf)^{-1}\Zbf\ybf_0 - \ybf_0 \big\|_2^2 = \|\ybf_0\|_2^2  - \big\|\Ubf(\Zbf)^{\intercal}\ybf_0 \big\|_2^2.
\end{equation}

Combining (\ref{eqn:generalization-eqn1}), (\ref{eqn:generalization-eqn2}) and (\ref{eqn:generalization-eqn3}), we have:
\begin{equation*}
\begin{split}
\Ebb_{\thetabf, \epsilonbf} \Big[ L\big(\Zbf\thetabf_2^*, \ybf_0\big) - L\big(\Xbf\thetabf_1^*, \ybf_0\big) \Big] & = \Ebb_{\thetabf}\big[\Lcal(\Zbf) - \Lcal(\Xbf)\big] \\ 
& \leq \Ebb_{\thetabf} \big[L(\Zbf\thetabf_2, \ybf_0) - L(\ybf_0, \ybf_0) + 2L\sigma \big] \\
& \leq \Ebb_{\thetabf}\Big[ \frac{L}{|\Ical|}\|\Zbf\thetabf_2 - \ybf_0\|_2 + 2L\sigma \Big] \quad (\text{using the Lipschitz condition of }L) \\
& \leq \Ebb_{\thetabf}\Big[ \frac{L}{|\Ical|}\sqrt{\|\ybf_0\|_2^2  - \big\|\Ubf(\Zbf)^{\intercal}\ybf_0 \big\|_2^2} + 2L\sigma \Big] \\
& \leq \frac{L}{|\Ical|} \sqrt{\Ebb_{\thetabf}\big[\|\ybf_0\|_2^2  - \big\|\Ubf(\Zbf)^{\intercal}\ybf_0 \big\|_2^2 \big]} + 2L\sigma \quad \text{(using Jensen's inequality)} \\
& \leq \frac{L}{|\Ical|} \sqrt{tr(\Sigmabf) - \underline{\lambda(\Sigmabf)}\big\| \Ubf(\Zbf)^{\intercal} \Ubf(\Xbf) \big\|_2^2} + 2L\sigma.
\end{split}
\end{equation*}
In the last line we use the definition of $\ybf_0$ to obtain: $\Ebb_{\thetabf}\Big[\big\|\Ubf(\Zbf)^{\intercal}\ybf_0 \big\|_2^2 \Big] = \big\|\Ubf(\Zbf)^{\intercal} \Ubf(\Xbf)\Sigmabf^{1/2}\big\|_F^2$, where $\|\cdot\|_F$ is the Frobenius norm. It is then easy to verify that $\big\|\Ubf(\Zbf)^{\intercal} \Ubf(\Xbf)\Sigmabf^{1/2}\big\|_F^2 \geq \underline{\lambda(\Sigmabf)}\big\| \Ubf(\Zbf)^{\intercal} \Ubf(\Xbf) \big\|_2^2$, which gives the desired result.
\end{proof}